\documentclass[conference]{IEEEtran}
\usepackage{graphicx} 
\usepackage{subfigure}
\usepackage{hyperref}
\hypersetup{colorlinks,allcolors=black}
\usepackage{multirow}
\usepackage{etoolbox}
\makeatletter
\patchcmd{\@makecaption}
  {\scshape}
  {}
  {}
  {}
\makeatletter
\patchcmd{\@makecaption}
  {\\}
  {.\ }
  {}
  {}
\makeatother

\ifCLASSINFOpdf
\else
\fi
\begin{document}
%
\title{Dual-Attention Model for Aspect-Level Sentiment Classification}

\author{\IEEEauthorblockN{Mengfei Ye}
\IEEEauthorblockA{School of Information Science and Engineering\\
East China University of Science and Technology\\
Shanghai, China\\
E-mail: Y80220116@mail.ecust.edu.cn}}


%


\maketitle

\begin{abstract}
I propose a novel dual-attention model(DAM) for aspect-level sentiment classification. Many methods have been proposed, such as support vector machines for artificial design features, long short-term memory networks based on attention mechanisms, and graph neural networks based on dependency parsing. While these methods all have decent performance, I think they all miss one important piece of syntactic information: dependency labels. Based on this idea, this paper proposes a model using dependency labels for the attention mechanism to do this task. We evaluate the proposed approach on three datasets: laptop and restaurant are from SemEval 2014, and the last one is a twitter dataset. Experimental results show that the dual attention model has good performance on all three datasets.
\end{abstract}


%
\IEEEpeerreviewmaketitle

\section{Introduction}
Natural language resources are usually presented in the form of text, behind which people's views on a certain commodity or event are often hidden. If we can accurately analyze the emotional attitude hidden behind the text, and use these data reasonably and efficiently, it will bring huge benefits to consumers, businesses, and relevant departments. Text-based sentiment analysis is an important task in the field of natural language processing. It can extract sentiments or viewpoints from a given text by analyzing its underlying sentiments. The task of sentiment analysis at document or sentence level is to conduct sentiment analysis on the whole document or sentence. However, because there are often multiple subjects with different emotional polarities in a document or sentence, such sentiment analysis method is often inaccurate. Therefore, more fine-grained aspect level sentiment analysis has attracted more and more attention\cite{poria2020beneath,hu2004mining}. The so-called aspect level sentiment analysis is to determine the emotional polarity of a particular aspect of the sentence, where the aspect refers to a particular thing or class of things. For example, in the sentence “I like coming back to Mac OS, but this laptop is lacking in speaker quality compared to my \$400 old HP laptop”, it is difficult to judge the whole sentiment polarity of this sentence. But in the aspect level sentiment classification, we can say that the polarity of the sentence towards the aspect “Mac OS” is positive while the polarity is negative in terms of the aspect “speaker quality”. In this way, we can get the exact emotional polarity of certain aspects which can be one hot event, one good, or the service, so the aspect level sentiment analysis is more accurate and useful than the coarse-grained sentiment analysis.

Review of existing ALSC methods, from the initial method based on hand-annotated feature engineering\cite{lin2011proceedings}, and then using deep learning, people started using various variants of LSTM to solve this problem\cite{tang2015effective}. With the introduction of attention\cite{wang2016attention} and more recently GNN\cite{sun2019aspect} to parse the grammatical features of dependency relationship, the model performance of aspect level sentiment analysis is getting stronger step by step. I think the performance of the aspect level sentiment analysis model has done a good job in semantic feature extraction. From the original LSTM\cite{hochreiter1997long} to the more powerful pre-training model BERT\cite{devlin2018bert}, I think the model has been difficult to improve in the simple semantic feature extraction. 

But in the extraction of grammatical features, I think there is still a lot of room for improvement. By observing the current models with a good performance involving the extraction of grammatical features, I find that these models simply extract connection information of dependency arc as grammatical features and input it into GCN\cite{daigavane2021understanding} or GAT\cite{velivckovic2017graph}. But in addition to the connection information of dependency arc, the label of dependency arc such as "direct object", "adjective modiﬁer" is also a very important part to provide syntactic information. So I think that the model can convert these dependency arc labels to embeddings to do the attention operation on the sentence to catch the important parts in this sentence. Also, considering the relationship between words and dependency arc labels in the dependency tree, we use the graph neural network to handle the word embedding and dependency label embedding.

\begin{figure}[h]
    \includegraphics[width=\linewidth]{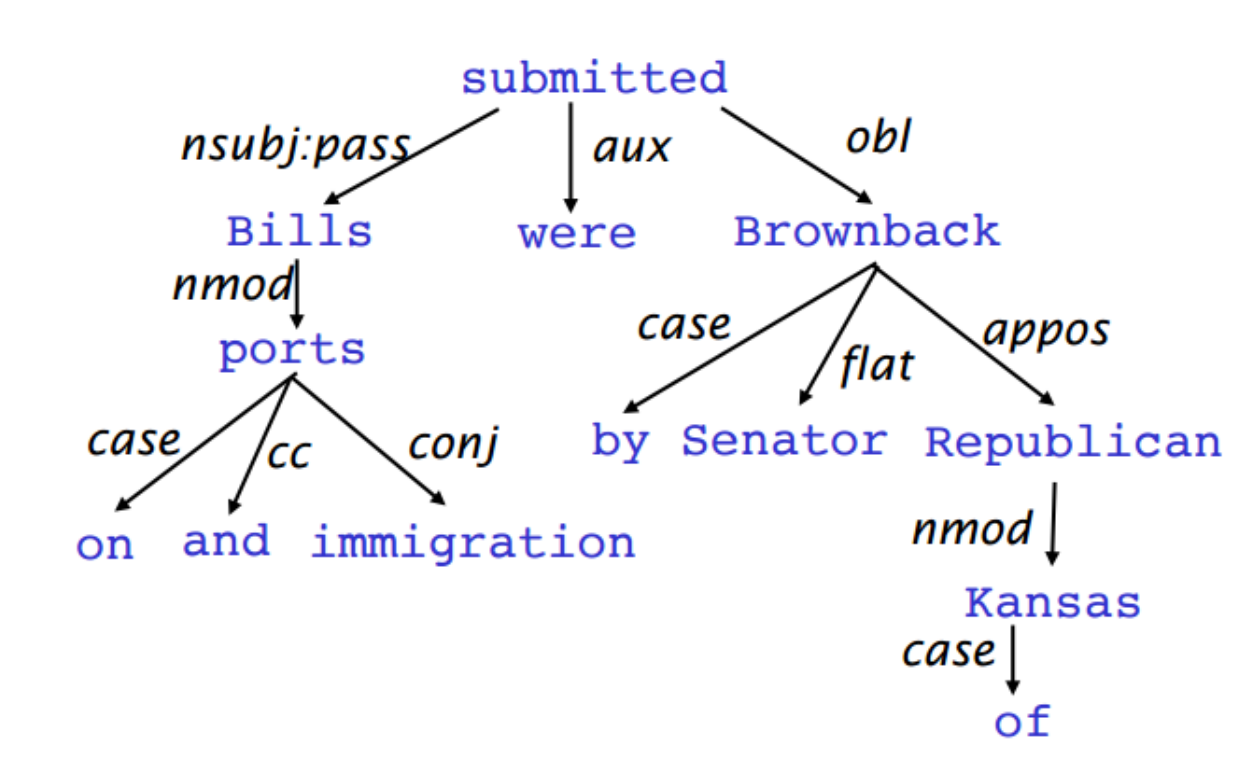}
    \centering
    \caption{an example of dependency tree}
\end{figure}

Therefore, based on my ideas mentioned above, I put forward two effective and powerful models to do the aspect level sentiment classification task in this paper. One is based on Bi-LSTM and the other is based on BERT. Both have relatively good performance. Although the model based on Bi-LSTM has a certain gap with the SOTA model, but the model based on BERT is comparable to the SOTA model's performance which is also based on BERT\cite{li2021dual}.
\section{Related Work}
\subsection{Attention mechanism}
Attention has become an effective mechanism for superior results, as demonstrated in image recognition, machine translation, reasoning about entailment, and sentence summarization\cite{wang2016attention}. Attention is the mechanism to focus on more important things but ignores irrelevant things. For example, in the sentence “The operating system of this laptop is good but the sound is bad.”, to do the aspect level sentiment analysis of operating, we should focus on the word “good” but ignore “bad”. But for the aspect of “sound”, we should focus on the word “bad” but ignore “good”.

Wang et al.\cite{wang2016attention} first proposes a novel approach for aspect-level sentiment classification using a combination of long short-term memory (LSTM) and attention mechanisms. Ma et al.\cite{ma2017interactive} is also to interact target and context to obtain accurate expression of sentences, using the model of attention. Different from the above models, the paper focuses on how to get a better vector representation of aspect words. As an improvement,  Fan et al.\cite{fan2018multi} proposes multi-grained attention network and aspect alignment loss. The multi-grained attention layer contains fine-grained attention layer and coarse-grained attention layer. Huang et al.\cite{huang2018aspect} first calculates the interaction matrix and then gets target-to-sentence attention and sentence-to-target attention. It averages the target-to-sentence attention to calculate target-level attention and finally calculates the sentence-level attention. 

Tang et al.\cite{tang2016aspect} uses memory network instead of LSTM to solve the ABSA problem. Song et al.\cite{song2019attentional} proposes attentional encoder network instead of RNN model with attention mechanism. The memory network maintains an external memory cell for storing previous information, rather than through hidden state inside the cell. And inspired by the combination of RNN encoder and attention mechanism decoder, Xing et al.\cite{xing2019earlier} proposes a novel variant of LSTM, termed as aspect-aware LSTM. They argue that aspect information should be considered in LSTM cells to improve the information flow. 

\subsection{Dependency Parsing}
To analyze the syntactic structure of sentences, the dependency tree is always used. The dependency tree of sentences shows which words depend on which other words so it can help the aspects find their contextual words. Generally, these dependency tree based ALSC models are implemented in two methods. The first one is to use the tree-based distance, which counts the number of edges in the shortest path between two tokens in the dependency tree. The second one is to use the topological structure of the dependency tree.

For the first one, Zhang et al.\cite{zhang2019syntax} propose a proximity-weighted convolution network to offer an aspect-specific syntax-aware representation of contexts. In this paper, two ways of determining proximity weight are explored, namely position proximity and dependency proximity. The position proximity is about the absolute position in the context, but the dependency proximity measures the distances between words in a syntax dependency parsing tree. And Nguyen et al.\cite{nguyen2018effective} introduces an attention model that incorporates syntactic information into the attention mechanism. They ignore context words whose location is larger than a certain window size and for context words within the window, different weights are applied so that words closer to the target receive more attention.

And for the second one, Zhang et al.\cite{zhang2019aspect} proposes to build a graph convolutional network over the dependency tree of a sentence to exploit syntactical information and word dependencies. It inputs the hidden states of the encoding layer with dependency parsing information into the graph convolutional network and does the aspect-specific masking of the output of GNN. Be more complex than the above paper, Liang et al.\cite{liang2020jointly} proposes a novel graph-aware model with interactive graph convolutional networks for aspect sentiment analysis. And Li et al.\cite{li2021dual} proposes the DualGCN model considering both the syntactic structure and the semantic correlation within a given sentence. This model integrates the SynGCN and SemGCN networks through a mutual BiAffine module. Instead of using graph convolutional network, Huang et al.\cite{huang2019syntax} use graph attention network\cite{velivckovic2017graph} to do this task.
\subsection{BERT}
BERT, which stands for bidirectional encoder representations from transformers is a powerful pretraining model\cite{devlin2018bert}. BERT is designed to pretrain deep bidirectional representations from the unlabeled text by joint conditioning on both the left and right context in all layers. As a result, the pre-trained BERT model can be fine-tuned with just one additional output layer to create state-of-the-art models for a wide range of tasks, such as question answering and language inference, without substantial task-specific architecture modifications. It obtains new state-of-the-art results on eleven natural language processing tasks in the authors’ experiments. In the aspect-level sentiment classification task, the BERT pretrain model also plays a powerful performance. In several datasets\cite{dong2014adaptive,pontiki2014semeval,jiang2019challenge} and models\cite{song2019attentional,liang2020jointly,huang2019syntax,jiang2019challenge}, if the BERT model replaces the encoding block of the original model, the accuracy can go up several percentages.
\section{Our Approach}
\subsection{Task Definition}
Given a sentence $s=\left\{w_1,w_2,\cdots{,w}_N\right\}$ consisting of $N$ words, and an aspect $a=\left\{w_{a1},w_{a2},\cdots{,w}_{am}\right\}$, the task of aspect level sentiment classification is to judge the sentiment polarity of the sentence s towards aspect $t$. Although the kinds of sentiment polarity are decided by the datasets, there are always three kinds of sentiment polarity which are positive, neutral, and negative. And the $w_i$ stands for the word in a sentence, and the $w_{ai}$ stands for the word of a certain aspect, the aspect can be made up of many words.

I present the overall architecture of the proposed Dual Attention Model(DAM) in figure-2. It consists of the Dependency Parsing Layer, the Embedding Layer, the Encoder Layer, the GCN Block Layer, the Attention Mechanism Layer, and the Final MLP Layer.

\begin{figure*}[h]
    \includegraphics[width=\linewidth]{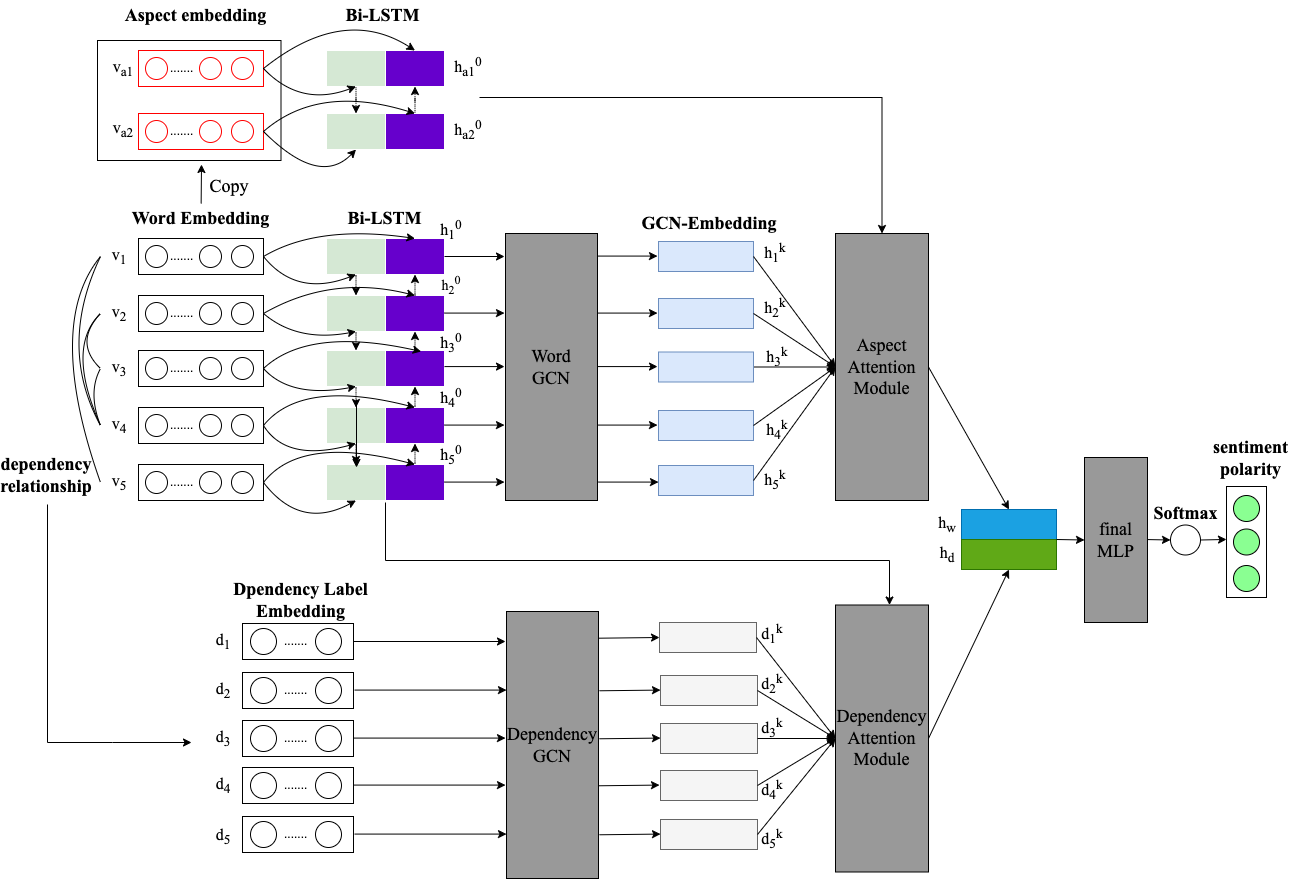}
    \centering
    \caption{Overall architecture of dual-attention model}
\end{figure*}

\subsection{Dependency Parsing Layer}
The dependency parsing Layer parses the dependency relationship of words in the sentence, and the dependency relationship includes the dependency arc and the label of the dependency arc. The dependency arc can also be understood as a connected relation between two words and this connection has the direction. Although in my processing way, I ignore the direction of the dependency arc and only use the connected relationship. Using the connection information provided by the dependency arc, the model will create a graph that is stored in one matrix G. And the graph is used in the word graph convolutional block and dependency graph convolutional block. For this matrix, if a sentence has n words, each word will depend on one word, so there will be n dependency arc labels and n dependency connection relation. To ignore the direction of the connection, if word $w_i$ depends on words $w_j$, both $G_{ij}$ and $G_{ji}$ will be filled which means the connection relation.
\subsection{Embedding Layer}
There are two kinds of the embedding layer in my model. The one is the word embedding layer which is used to convert certain words to a specific word vector. The other one is the dependency label embedding layer which is used to convert certain labels of dependency relationships to a specific vector representation.

In this word embedding layer, I use Glove embedding to convert words to word vectors. I construct a word vector matrix $M^{V\times{dim}_w}$ where the V represents the number of kinds of words in the dataset, and ${dim}_w$ represents the dimension of the word vector. I initialize the word vector matrix using glove embedding finally we can get the word vectors $\left\{v_1,v_2,{\cdots,v}_n\right\}$ by multiplying the inputs with word vector matrix where $v_i\in R^{{dim}_w}$.

The dependency embedding label layer is like the word embedding layer. To convert the dependency label to a certain vector representation, the model will also construct a matrix $M^{V^{\prime}\times{dim}_l}$ where the $V^{\prime}$ is the number of kinds of dependency label in the dependency result about the specific dataset, and ${dim}_l$ represents the dimension of the vector representation. This layer outputs the vector representation about dependency arc labels $\left\{d_1,d_2,{\cdots,d}_n\right\}$ where $d_i\in R^{{dim}_l}$. 

\subsection{Encoder Layer}
To learn more sufficient semantic information, I use the bi-LSTM network as the encoder. In my model, there are two parts of Bi-LSTM that handle the sentence information and aspect information respectively.

For the sentence handling module, the model inputs the sentence vector $\left\{v_1,v_2,{\cdots,v}_n\right\}\in R^{n\times{dim}_w}$ into the forward LSTM to obtain the forward hidden layer output $\vec{h_s}\in R^{n\times d_h}$. At the same time, the sentence vector is input into the reverse LSTM to obtain the reverse hidden layer output $\mathop{h_s}\limits ^{\leftarrow}\in R^{n\times d_h}$, finally concatenates $\vec{h_s}$ and $\mathop{h_s}\limits ^{\leftarrow}$ to get the result $h_s\in R^{n\times{2d}_h}$. Meanwhile, the aspect encoding process is similar, $h_a\in R^{m\times{2d}_h}$ can be obtained after bi-LSTM. The $h_s$ is the gathering of $\left[h_1^0;h_2^0;\cdots;h_n^0\right]$ and the $h_a$ is the gathering of $\left[h_{a_1}^0;h_{a_2}^0;\cdots;h_{a_m}^0\right]$ corresponds to the overall architecture figure.







\subsection{GCN Block Layer}
In the previous dependency parsing layer, through the connection information of the dependency relationship, we can get the connection graph $G\in R^{n\times n}$ where n is the number of words in this sentence, the item in this matrix is either $1$ or $0$ to stand for connect or not between two words. In this graph convolutional network block, the model will use the graph and feature vectors to do special computing. There are two parts in the GCN block which are the word GCN and the dependency label GCN.

In the word GCN block, the model inputs the $h_s\in R^{n\times{2d}_h}$ and $G\in R^{n\times n}$ into the graph convolutional network. And the output of the word GCN block is $h_s^k\in R^{n\times{dim}_{wordgcn}}$ where the $k$ means the number of words GCN’s layer and ${dim}_{wordgcn}$ means the output dimension of final word GCN layer. The $h_s^k$ is the gathering of $\left[h_1^k;h_2^k;\cdots;h_n^k\right]$ corresponds to the overall architecture figure. The detail of the graph convolutional network is shown in the following equation. In this formula, $l$ means the layer, $H$ is the hidden state of the whole graph, $\sigma$ is the activation function, $W$ is the parameters of the neural network, ${\hat{D}}$ is the degree matrix to do normalization, $\hat{A}$ is the sum of the graph adjacency matrix $A$ and the diagonal matrix $I$. In rough, the graph convolutional network can process the connection relation about nodes in the graph. If the word $w_i$ depends on the word $w_j$, the word $w_i$ should fuse the word $w_j$’s information into self’s vector, the graph convolutional network is a powerful technology to do this thing. And, using the GCN with a deep layer the features of words can be broadcast widely.

\begin{equation}
    H^{\left(l+1\right)}=\sigma\left({\hat{D}}^{-\frac{1}{2}}\hat{A}{\hat{D}}^{-\frac{1}{2}}H^{\left(l\right)}W^{\left(l\right)}\right)
\end{equation}

The dependency label GCN block is like the word GCN block, but the input is the vectors about dependency labels instead of words. In the word GCN block, the model inputs the $D\in R^{n\times{dim}_l}$ and $G\in R^{n\times n}$ into the graph convolutional network in which the D is the aggregation of $\left\{d_1,d_2,{\cdots,d}_n\right\}$ where $d_i\in R^{{dim}_l}$. The output of the dependency label GCN block is $D^k\in R^{n\times{dim}_{depgcn}}$ where the $k$ means the number of dependencies GCN’s layer and ${dim}_{depgcn}$ means the output dimension of final dependency label GCN layer. About the dependency relationship, if $w_i$ depends on $w_j$ and $w_j$ depends on $w_k$, the $w_i$ also depends on the $w_k$. The dependency label vectors are used to do the attention on sentence vectors, with the single link between $w_i$ and $w_j$ or $w_j$ and $w_k$, the model can not grasp the underlying meaning. But with the graph convolutional network, this deep-level dependency label information can be mined.

\subsection{Attention Mechanism Layer}
The attention mechanism module is the most important part of my Dual-Attention model. There are two kinds of attention modules in this model, the aspect attention module and the dependency attention module. These two modules pay attention to different parts. The aspect attention module regards aspect embedding as query and sentence feature vectors as key and value for attention, which focuses on finding features that have a strong relationship with a particular aspect. While the dependency attention module takes the dependency label feature processed by graph convolutional network as query, the sentence feature vectors as key and value for attention, which focuses on finding some words in a sentence that can best represent emotional polarity. In my Dual-Attention model, I use aspect attention and dependency attention in parallel to reason emotional polarity and finally concatenate the feature vectors after the two attention mechanisms into the final MLP for classification.

The aspect attention module uses the dot product attention mechanism. The query for this attention mechanism is aspect embedding $h_a\in R^{m\times{2d}_h}$. The key and value of this attention mechanism is the output of word graph convolutional network $h_s^k\in R^{n\times{dim}_{wordgcn}}$. In the setting of my model, the value of ${dim}_{wordgcn}$ is equal to the value of ${2\times d}_h$. So, in the aspect attention module, I use the dot product attention as the following equation. And the output of this aspect attention module is $h_w\in R^{{2d}_h}$.

\begin{equation}
    aspect-att. \left(\ h_a,h_s^k,h_s^k\right)=softmax\left(\frac{\ h_a{{(h}_s^k)}^T}{\sqrt{{2d}_h}}\right)h_s^k
\end{equation}

Different from the aspect attention module, the dependency attention module uses the additive attention mechanism. The query for this attention mechanism is the output of dependency label graph convolutional network $D^k\in R^{n\times{dim}_{depgcn}}$. The key and value of this attention mechanism are the hidden states of words outputted by Bi-LSTM $h_s\in R^{n\times{2d}_h}$. So, in the dependency attention module, I use additive attention as the following equation. The $W_q$, $W_k$, and $W_v$ are the learnable parameters. And the output of this dependency attention module is $h_d\in R^{{2d}_h}$.

\begin{equation}
    dp-att. \left(D^k,h_s,h_s\right)=w_v^Ttanh(W_qD^k+W_kh_s)h_s
\end{equation}

\subsection{Final MLP Layer}
The final multilayer perceptron is several simple fully connected layers to do the classification. In the aspect attention module and dependency attention module, we get the result $h_w\in R^{{2d}_h}$ and $h_d\in R^{{2d}_h}$. To do the final classification, we concentrate them to be a new feature vector $h^\ast\in R^{{4d}_h}$ which is the input of the final MLP. In the final multilayer perceptron, we can get the output $y^\prime\in R^3$.

\begin{equation}
    y^\prime=relu(W_2·relu(W_1h^*+b1)+b2)
\end{equation}

And the SoftMax layer is needed for sentiment classification.

\begin{equation}
    \hat{y}=softmax(y^\prime)
\end{equation}

\subsection{Bert-based Model}
The Bert-based model uses the pre-trained BERT model as an encoder replacing Bi-LSTM which brings huge performance improvement in accuracy and f1 score.

The main idea of this Bert-based model is same as the original dual-attention model. The model structure from dependency label embedding Layer to dependency attention module in the figure is the same as the original model. The difference between Bert-based model and original model is mainly in the upper part of the figure.  The input of BERT model is sentence $\left\{w_1,w_2,\cdots,w_n\right\}$, these symbols represent the strings but not the word vectors. To change the words into the form that BERT model can process, we first use BERT tokenizer to divide each word into tokens. Note that this token is usually not the whole word, but the root or affixes that make up word.  In processing, we have tokenized the whole sentence and the aspect respectively, so we get two parts results. We then combine the two results, add the [SEP] delimiter between the two results, add the [CLS] delimiter on the head and add the [SEP] delimiter on the tail as shown in figure-3. Then we have one input for Bert model. Another input is the mask that distinguishes the two segmented results. After the two inputs are inputted into the pretrained BERT model, and the BERT model processes the inputs by its powerful bi-directional transformer structure, two outputs are obtained. One is the pooled vector $v_{pool}\in R^{dim}$ where the dim is the hidden size of BERT model. Another is the hidden state $h_{final}\in R^{m\times dim}$ of the sentence tokenized input in the last layer of the model.  

The pooled vector $v_{pool}\in R^{dim}$ is the feature vector that can be inputted into the final MLP directly for sentiment classification to get a relatively good result. But doing this will make the model a pure BERT model. To make use of the rich syntactic information, we use the final hidden state $h_{final}\in R^{m\times dim}$ to do the dependency attention. Because of the unique tokenizing way of Bert, the first dimension of hidden state $h_{final}\in R^{m\times dim}$ is m but not the words’ number n of sentence. So, I first make the vector back to the length of the original length and then I get the hidden state vector $h_{final}^\prime\in R^{n\times dim}$. And then we do the dependency attention which is the same as the original model. Finally, we concatenate the $v_{pool}$ and $h_{final}^\prime$ vector to get one feature vector. And we input the feature vector to the final MLP layer to get the sentiment polarity. By the way, the $v_{pool}$ and $h_{final}^\prime$ vector in the figure-3 are $h_w$ and $h_d$ respectively.

\begin{figure*}[h]
    \includegraphics[width=\linewidth]{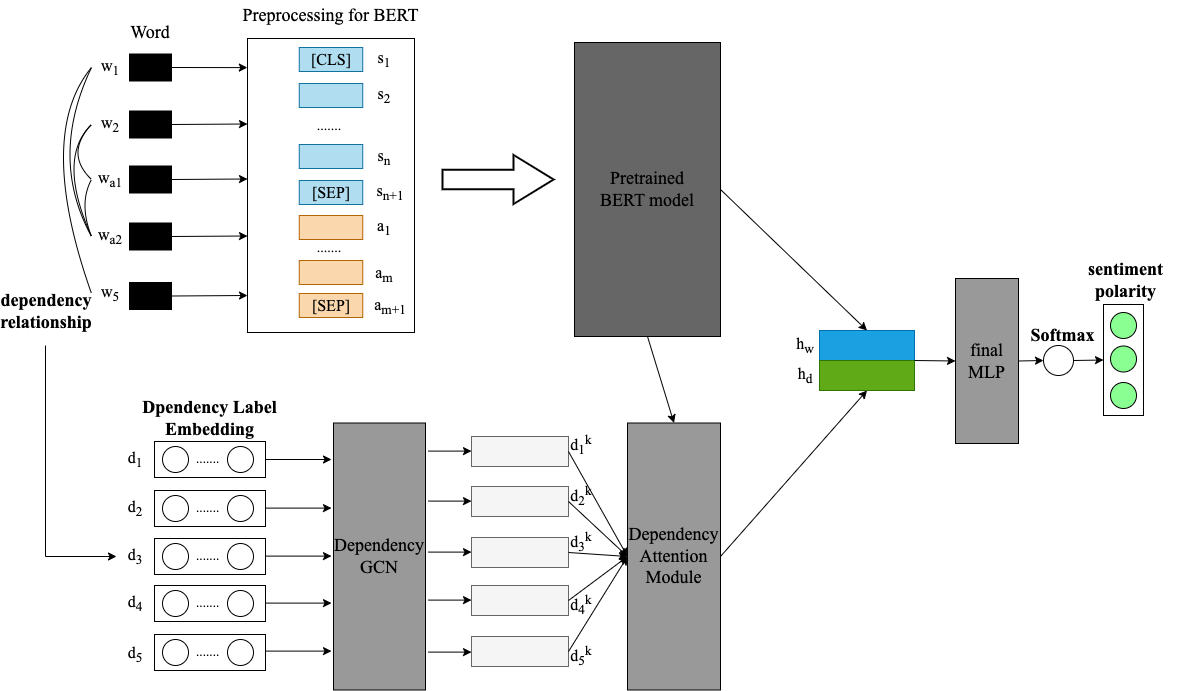}
    \centering
    \caption{Overall architecture of Bert-based model}
\end{figure*}
\section{Experiments}
\subsection{Experiment Setting}
In the experiments, I use three datasets to evaluate my model’s performance, as shown in following table. The first two are from the SemEval 2014 Task 4\cite{pontiki2014semeval}, which contains the reviews in laptop and restaurants, respectively. The third one is a tweet collection, which are gathered by paper\cite{dong2014adaptive}. Each aspect with the context is labeled by three sentiment polarities, namely positive, neutral, and negative.

We use the cross-entropy loss as the objective function to train my model. The $\hat{y}\in R^3$ is the predicted sentiment distribution and $y\in R^3$ is the ground truth distribution. $\lambda$ is the L2 regularization term and $\theta$ is the parameter set. \footnote{More details can be found in GitHub repo: \href{https://github.com/mufiye/mufiye_ALSC}{https://github.com/mufiye/mufiye\_ALSC}.}

\begin{equation}
\mathcal{J}(\theta)=-\sum_{i}\sum_{j}{y_i^j\log{{\hat{y}}_i^j}}+\lambda||\theta||^{2}
\end{equation}

In my original model’s training process, I use the Adam optimizer\cite{duchi2011adaptive} to optimize the training process. And for the Bert-based model’s training process, I use the AdamW optimizer\cite{loshchilov2017decoupled} instead.

To evaluate the performance of the model, we adopt two evaluation metrics: accuracy and F1-score. The metrics are widely used in previous works\cite{wang2016attention,ma2017interactive,fan2018multi}.

All experiments use the Pytorch framework with Cuda and run on one NVIDIA 3090 GPU. The embedding size of glove word embedding is 300. The dependency label embedding dimension is set to 300 and the hidden size of Bi-LSTM is 100. The pretrained Bert model's hidden size is 768. I set the start learning rate ${10}^{-3}$ to the original model and $5^{-5}$ to the Bert-based model. The training batch size is set to 32. The adam epsilon of AdamW optimizer is set to ${10}^{-8}$. In addition to these parameters, other parameters in models are initialized by sampling from a uniform distribution $U(-\epsilon,\ \epsilon)$.
\subsection{Overall Performance Comparison}
I compare my ALSC models with state-of-the-art baselines. These baseline models are briefly described as follows. And the comparison result is shown in the following table-1.

(1)ATAE-LSTM\cite{wang2016attention} utilizes aspect embedding and the attention mechanism in aspect-level sentiment classification.

(2)IAN\cite{ma2017interactive} employs two LSTMs and an interactive attention mechanism to generate representations for the aspect and sentence.

(3)RAM\cite{chen2017recurrent} uses multiple attention and memory networks to learn sentence representation.

(4)AEN\cite{song2019attentional} designs an attentional encoder network to draw the hidden states and semantic interactions between the target and context words.

(5)ASGCN\cite{zhang2019aspect} first proposed using GCN to learn the aspect-specific representations for aspect-based sentiment classification.

(6)kumaGCN\cite{chen2020inducing} employs a latent graph structure to complement syntactic features.

(7)R-GAT\cite{wang2020relational} proposes an aspect-oriented dependency tree structure and then encodes new dependency trees with a relational GAT.

(8)DualGCN\cite{li2021dual} uses SynGCN and SemGCN networks through a mutual BiAffine module for aspect-level sentiment classification tasks. This model is the state-of-the-art model in ALSC tasks as far as know. 

(9)Pure BERT\cite{devlin2018bert} is the vanilla BERT model by feeding the sentence-aspect pair and using the representation of [CLS] for predictions.

\begin{table*}[]
\centering
\begin{tabular}{ccccccc}
\hline
\multirow{2}{*}{\textbf{Models}} & \multicolumn{2}{c}{\textbf{Restaurant}}     & \multicolumn{2}{c}{\textbf{Laptop}}         & \multicolumn{2}{c}{\textbf{Twitter}}        \\ \cline{2-7} 
                                 & \textbf{Accuracy}    & \textbf{Macro-F1}    & \textbf{Accuracy}    & \textbf{Macro-F1}    & \textbf{Accuracy}    & \textbf{Macro-F1}    \\ \hline
ATAE-LSTM                        & 77.2                 & -                    & 68.7                 & -                    & -                    & -                    \\
IAN                              & 78.6                 & -                    & 72.1                 & -                    & -                    & -                    \\
RAM                              & 80.23                & 70.8                 & 74.4                 & 71.35                & 69.36                & 67.3                 \\
AEN                              & 80.98                & 72.14                & 73.51                & 69.04                & 72.83                & 69.81                \\
ASGCN                            & 80.77                & 72.02                & 75.55                & 71.05                & 72.15                & 70.4                 \\
KumaGCN                          & 81.43                & 73.64                & 76.12                & 72.42                & 72.45                & 70.77                \\
R-GAT                            & 83.3                 & 76.08                & 77.42                & 73.76                & 75.57                & 73.82                \\
DualGCN                          & \textbf{84.27}       & \textbf{78.08}       & \textbf{78.48}       & \textbf{74.74}       & \textbf{75.92}       & \textbf{74.29}       \\ \hline
\textbf{My dual-att. model}      & \textbf{81.25}       & \textbf{72.53}       & \textbf{75.39}       & \textbf{71.16}       & \textbf{73.12}       & \textbf{72.05}       \\ \hline
Pure Bert                        & 85.09                & 77.88                & 77.74                & 73.71                & 74.71                & 73.54                \\
AEN-BERT                         & 83.12                & 73.76                & 79.93                & 76.31                & 74.71                & 73.13                \\
R-GAT+BERT                       & 86.6                 & 81.35                & 78.21                & 74.07                & 76.15                & 74.88                \\
DualGCN+BERT                     & \textbf{87.13}       & 81.16                & \textbf{81.8}        & \textbf{78.1}        & \textbf{77.4}        & \textbf{76.02}       \\ \hline
\textbf{My Bert-based model}     & \textbf{86.96}       & \textbf{81.58}       & \textbf{80.88}       & \textbf{76.94}       & \textbf{76.3}        & \textbf{74.82}       \\ \hline
\multicolumn{1}{l}{}             & \multicolumn{1}{l}{} & \multicolumn{1}{l}{} & \multicolumn{1}{l}{} & \multicolumn{1}{l}{} & \multicolumn{1}{l}{} & \multicolumn{1}{l}{}
\end{tabular}
\caption{Experimental results comparison on three publicly available datasets.}
\label{tab:default}
\centering
\end{table*}

For the comparison of performance, as you can see, my dual-attention model thoroughly defeats the model which is based on attention mechanisms such as ATAE-LSTM\cite{wang2016attention}, IAN\cite{ma2017interactive}, RAM\cite{chen2017recurrent}, and AEN\cite{song2019attentional}. Compared with the early networks using graph neural networks, such as ASGCN\cite{zhang2019aspect} and kumaGCN\cite{chen2020inducing}, we are not inferior and even surpass in the performance of some datasets. The dual-attention model’s accuracy in the restaurant dataset is better than the ASGCN\cite{zhang2019aspect}. And the dual-attention model’s accuracy and f1 score in the Twitter dataset is better than the ASGCN\cite{zhang2019aspect} and kumaGCN\cite{chen2020inducing}. Although the performance of our dual-attention model has a certain gap with the top models\cite{li2021dual,wang2020relational} of these two years. 

But if you focus on the Bert-based model, you can find that my model has a wonderful performance in the top models of these two years. My Bert-based model’s performance is better pure-Bert\cite{devlin2018bert} which proves the validity of my dependency attention module from the side. My Bert-based model’s performance is better than the AEN-BERT\cite{song2019attentional}, R-GAT+BERT\cite{wang2020relational}, and close to the SOTA model DualGCN\cite{li2021dual}. 

From the above comparison analysis, I think that the model designed by me is useful for solving aspect-level sentiment classification tasks. Also, I think these two models have increasing space because of my poor training process.

\subsection{Analysis of DAM}
\subsubsection{My Dual-Attention Model vs Non-Dep Model}
To observe the effectiveness and necessity of dependency attention, I remove the dependency embedding layer, dependency GCN and dependency attention module to create a new model named the non-dep model. And I try to train this model, finally, I get the result shown in the following table-2. As you can see, the model with no dependency module has a gap with our original model in all three public datasets. This shows that dependency attention is helpful for improving the performance of the model.
\begin{table*}[]
\centering
\begin{tabular}{ccccccc}
\hline
\multirow{2}{*}{\textbf{Models}} & \multicolumn{2}{c}{\textbf{Restaurant}}     & \multicolumn{2}{c}{\textbf{Laptop}}         & \multicolumn{2}{c}{\textbf{Twitter}}        \\ \cline{2-7} 
                                 & \textbf{Accuracy}    & \textbf{Macro-F1}    & \textbf{Accuracy}    & \textbf{Macro-F1}    & \textbf{Accuracy}    & \textbf{Macro-F1}    \\ \hline
Non-dep model                    & 79.37                & 68.68                & 73.67                & 68.29                & 71.1                 & 69.05                \\
\textbf{My dual-att. model}      & \textbf{81.25}       & \textbf{72.53}       & \textbf{75.39}       & \textbf{71.16}       & \textbf{73.12}       & \textbf{72.05}       \\ \hline
\multicolumn{1}{l}{}             & \multicolumn{1}{l}{} & \multicolumn{1}{l}{} & \multicolumn{1}{l}{} & \multicolumn{1}{l}{} & \multicolumn{1}{l}{} & \multicolumn{1}{l}{}
\end{tabular}
\caption{my dual-attention model compared with non-dep model. The non-dep model means that my dual-attention model deletes the dependency embedding, dependency GCN and dependency attention module.}
\label{tab:default}
\centering
\end{table*}

\subsubsection{Other Comparison Results}
In the model-building process, I try to do the aspect attention and dependency attention serially without parallelly doing the attention mechanism and then concatenating them. In this model, the other layers work as usual except for the dependency attention module and aspect attention module. After the word GCN, the dependency features are processed by the dependency embedding layer, dependency GCN layer is used to do the dependency attention. And the result of dependency attention is inputted into the aspect attention module to get the final representation vector which is inputted into the final MLP to do classification. But as you can see in the following table-3, the performance of this model is not particularly ideal. My dual-attention model is better than this model. So finally, I choose the model which does parallel attention and concatenates the result vectors as my design.
\begin{table*}[]
\centering
\begin{tabular}{ccccccc}
\hline
\multirow{2}{*}{\textbf{Models}} & \multicolumn{2}{c}{\textbf{Restaurant}}     & \multicolumn{2}{c}{\textbf{Laptop}}         & \multicolumn{2}{c}{\textbf{Twitter}}        \\ \cline{2-7} 
                                 & \textbf{Accuracy}    & \textbf{Macro-F1}    & \textbf{Accuracy}    & \textbf{Macro-F1}    & \textbf{Accuracy}    & \textbf{Macro-F1}    \\ \hline
Another designed model           & 80.09                & 68.79                & 72.73                & 67.8                 & 72.25                & 69.89                \\
\textbf{My dual-att. model}      & \textbf{81.25}       & \textbf{72.53}       & \textbf{75.39}       & \textbf{71.16}       & \textbf{73.12}       & \textbf{72.05}       \\ \hline
\multicolumn{1}{l}{}             & \multicolumn{1}{l}{} & \multicolumn{1}{l}{} & \multicolumn{1}{l}{} & \multicolumn{1}{l}{} & \multicolumn{1}{l}{} & \multicolumn{1}{l}{}
\end{tabular}
\caption{my dual-attention model compares to the dual-attention model with another design.}
\label{tab:default}
\centering
\end{table*}
\subsection{Case Study}
To study my model’s inferring process and the roles of two attention modules, I do the case study on the sentence “Although the food was outstanding, but the little perks were bad.”. In this sentence, there are two aspects to analyze, the sentiment polarity of food is positive, and the sentiment polarity of perks is negative. The aspect attention module and dependency attention module are the most important modules in my dual-attention model. So, I visualize the attention weights of aspect attention and dependency attention. You can see the results in the figure-4 and figure-5. In the “food” aspect’s attention module, the attention weights of “the” and “food” are large, but for the “perks” aspect, the attention weights of the latter part of a sentence are large. And in the dependency attention module, you can find the attention weights for the aspect “food” and aspect “perks” are the same. And the dependency attention module focuses on words that have rich emotional information such as “outstanding”, “little” and “bad”.

\begin{figure}[htbp]
\centering
\subfigure[]{
\begin{minipage}[t]{0.5\linewidth}
\centering
\includegraphics[width=1.5in]{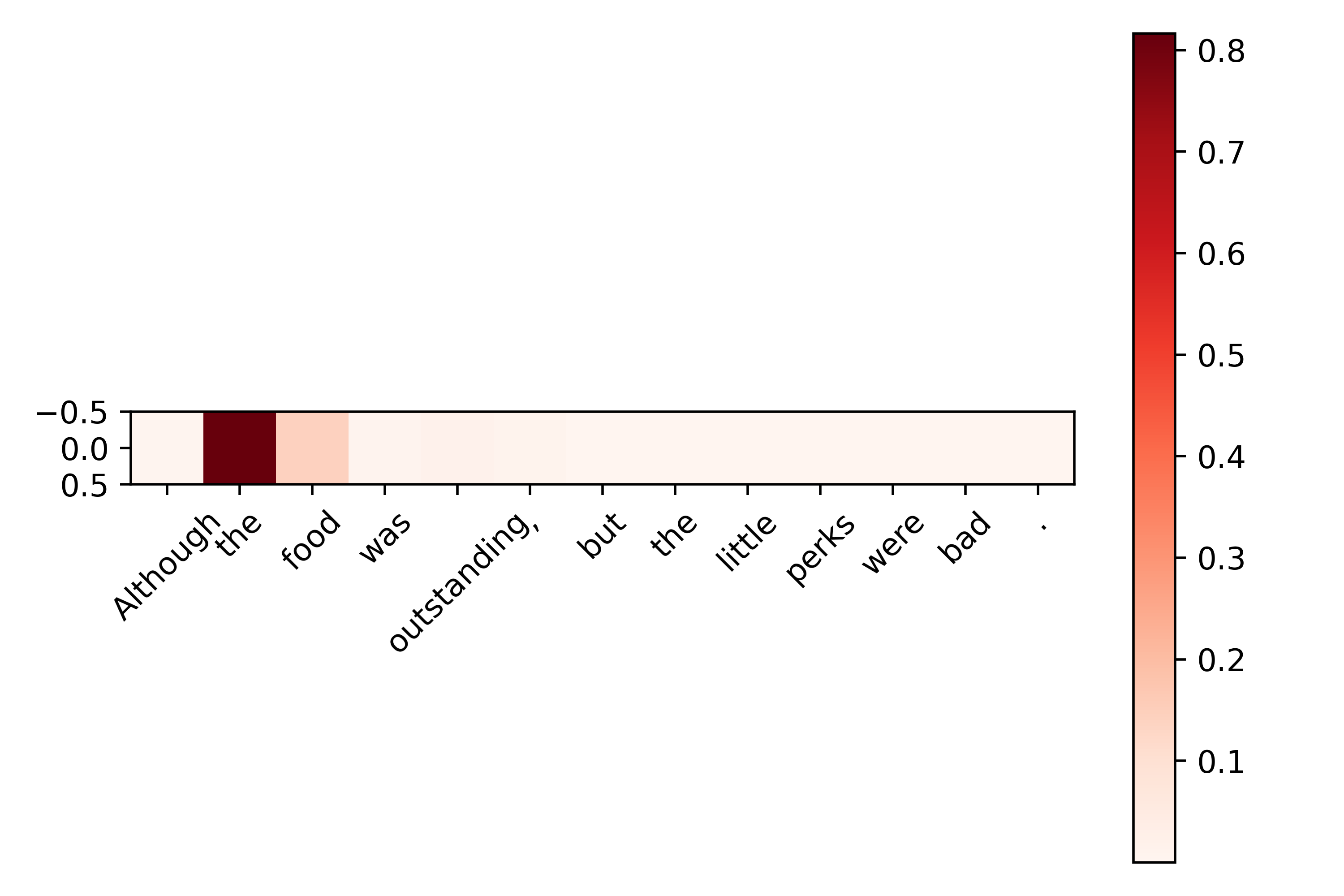}
\end{minipage}%
}%
\subfigure[]{
\begin{minipage}[t]{0.5\linewidth}
\centering
\includegraphics[width=1.5in]{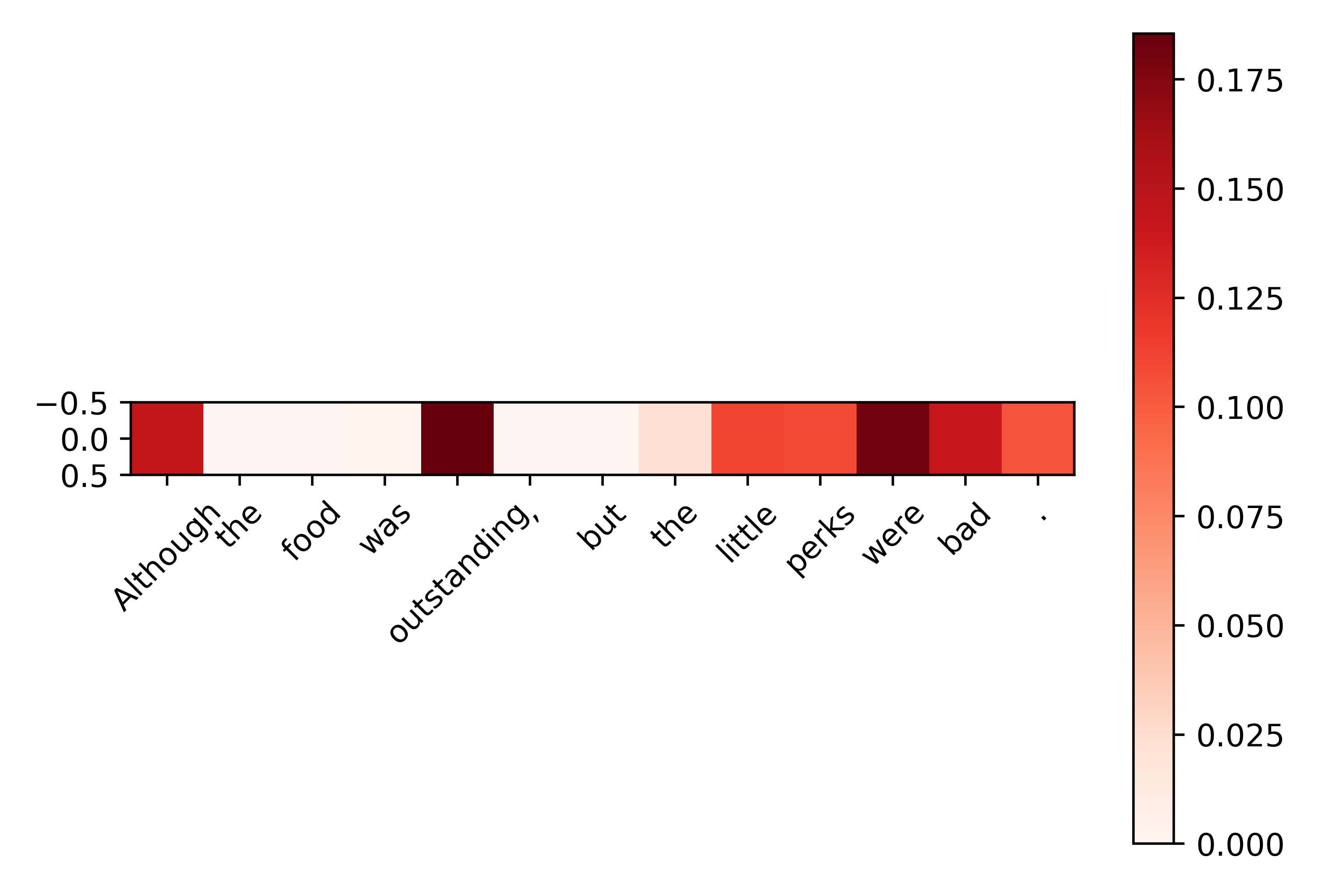}
\end{minipage}%
}%
\centering
\caption{attention heatmap of “food” aspect, left figure is weight of aspect attention, right figure is weight of dependency attention}
\end{figure}

\begin{figure}[htbp]
\centering
\subfigure[]{
\begin{minipage}[t]{0.5\linewidth}
\centering
\includegraphics[width=1.5in]{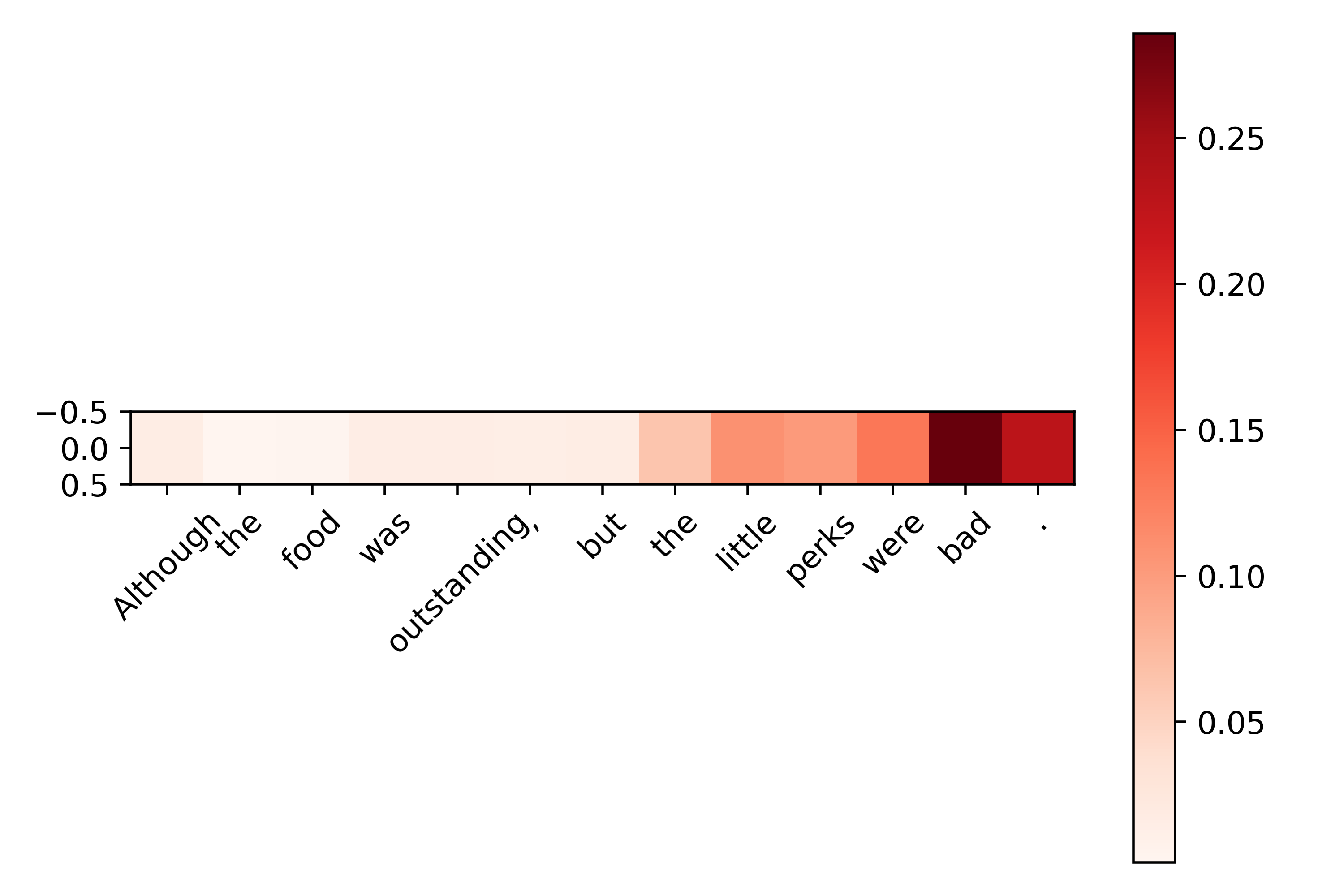}
\end{minipage}%
}%
\subfigure[]{
\begin{minipage}[t]{0.5\linewidth}
\centering
\includegraphics[width=1.5in]{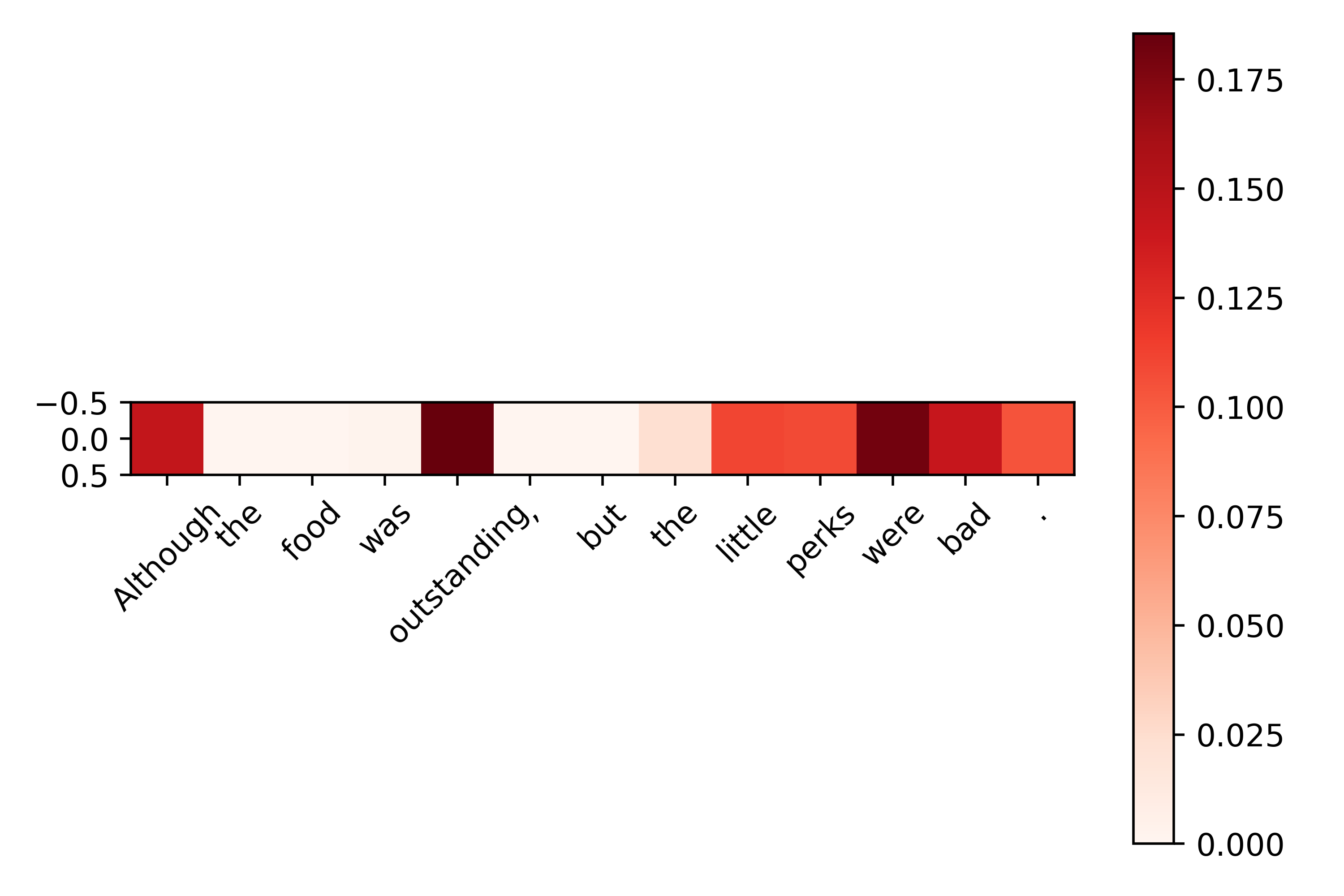}
\end{minipage}%
}%
\centering
\caption{attention heatmap of “perks” aspect, left figure is the weight of aspect attention, right figure is the weight of dependency attention.}
\end{figure}

By analyzing the attention weight on the word position, we can find that the aspect attention module always makes the model pay attention to the features which are close to the aspect. And the dependency attention module always makes the model pay attention to the features which have rich emotional information.

\section{Conclusion}
In this paper, I propose models with both an aspect attention module and a dependency attention module to solve the aspect-level sentiment classification task. The dependency attention module embeds the dependency label information and does an attention mechanism on word features to capture the words with a strong emotional tendency. Experiments show that the new proposed dependency attention module is effective and is the right design. And to enhance the performance of the original model, I use the pre-trained Bert model to replace the glove embedding layer and Bi-LSTM. The Bert-based model shows excellent performance on three public datasets. 

The performance of my dual-attention model still has a gap with the SOTA model. I think part of the reason for this is that I haven't found the best model parameters and training parameters for the model I designed. Because of the device, I just tried out a few possible sets of parameters on my model. I will try to train the performance of this model better in the future. At the same time, another part of the reason for this gap is that some networks in the model are not perfectly designed, resulting in poor convergence to the global minimum of neural networks. I think how to use dependency parsing arc and label is worth exploring in depth.


\section*{Acknowledgment}
I would like to thank the open-source researchers who worked on this aspect-level sentiment classification before because your code made it possible for me to conduct my research and experiments more efficiently.



%
\bibliographystyle{IEEEtran}
\bibliography{latex/main.bib}

\begin{thebibliography}{10}
\providecommand{\url}[1]{#1}
\csname url@samestyle\endcsname
\providecommand{\newblock}{\relax}
\providecommand{\bibinfo}[2]{#2}
\providecommand{\BIBentrySTDinterwordspacing}{\spaceskip=0pt\relax}
\providecommand{\BIBentryALTinterwordstretchfactor}{4}
\providecommand{\BIBentryALTinterwordspacing}{\spaceskip=\fontdimen2\font plus
\BIBentryALTinterwordstretchfactor\fontdimen3\font minus
  \fontdimen4\font\relax}
\providecommand{\BIBforeignlanguage}[2]{{%
\expandafter\ifx\csname l@#1\endcsname\relax
\typeout{** WARNING: IEEEtran.bst: No hyphenation pattern has been}%
\typeout{** loaded for the language `#1'. Using the pattern for}%
\typeout{** the default language instead.}%
\else
\language=\csname l@#1\endcsname
\fi
#2}}
\providecommand{\BIBdecl}{\relax}
\BIBdecl

\bibitem{poria2020beneath}
S.~Poria, D.~Hazarika, N.~Majumder, and R.~Mihalcea, ``Beneath the tip of the
  iceberg: Current challenges and new directions in sentiment analysis
  research,'' \emph{IEEE Transactions on Affective Computing}, 2020.

\bibitem{hu2004mining}
M.~Hu and B.~Liu, ``Mining and summarizing customer reviews,'' in
  \emph{Proceedings of the tenth ACM SIGKDD international conference on
  Knowledge discovery and data mining}, 2004, pp. 168--177.

\bibitem{lin2011proceedings}
D.~Lin, Y.~Matsumoto, and R.~Mihalcea, ``Proceedings of the 49th annual meeting
  of the association for computational linguistics: Human language
  technologies,'' in \emph{Proceedings of the 49th Annual Meeting of the
  Association for Computational Linguistics: Human Language Technologies},
  2011.

\bibitem{tang2015effective}
D.~Tang, B.~Qin, X.~Feng, and T.~Liu, ``Effective lstms for target-dependent
  sentiment classification,'' \emph{arXiv preprint arXiv:1512.01100}, 2015.

\bibitem{wang2016attention}
Y.~Wang, M.~Huang, X.~Zhu, and L.~Zhao, ``Attention-based lstm for aspect-level
  sentiment classification,'' in \emph{Proceedings of the 2016 conference on
  empirical methods in natural language processing}, 2016, pp. 606--615.

\bibitem{sun2019aspect}
K.~Sun, R.~Zhang, S.~Mensah, Y.~Mao, and X.~Liu, ``Aspect-level sentiment
  analysis via convolution over dependency tree,'' in \emph{Proceedings of the
  2019 conference on empirical methods in natural language processing and the
  9th international joint conference on natural language processing
  (EMNLP-IJCNLP)}, 2019, pp. 5679--5688.

\bibitem{hochreiter1997long}
S.~Hochreiter and J.~Schmidhuber, ``Long short-term memory,'' \emph{Neural
  computation}, vol.~9, no.~8, pp. 1735--1780, 1997.

\bibitem{devlin2018bert}
J.~Devlin, M.-W. Chang, K.~Lee, and K.~Toutanova, ``Bert: Pre-training of deep
  bidirectional transformers for language understanding,'' \emph{arXiv preprint
  arXiv:1810.04805}, 2018.

\bibitem{daigavane2021understanding}
A.~Daigavane, B.~Ravindran, and G.~Aggarwal, ``Understanding convolutions on
  graphs,'' \emph{Distill}, vol.~6, no.~9, p. e32, 2021.

\bibitem{velivckovic2017graph}
P.~Veli{\v{c}}kovi{\'c}, G.~Cucurull, A.~Casanova, A.~Romero, P.~Lio, and
  Y.~Bengio, ``Graph attention networks,'' \emph{arXiv preprint
  arXiv:1710.10903}, 2017.

\bibitem{li2021dual}
R.~Li, H.~Chen, F.~Feng, Z.~Ma, X.~Wang, and E.~Hovy, ``Dual graph
  convolutional networks for aspect-based sentiment analysis,'' in
  \emph{Proceedings of the 59th Annual Meeting of the Association for
  Computational Linguistics and the 11th International Joint Conference on
  Natural Language Processing (Volume 1: Long Papers)}, 2021, pp. 6319--6329.

\bibitem{ma2017interactive}
D.~Ma, S.~Li, X.~Zhang, and H.~Wang, ``Interactive attention networks for
  aspect-level sentiment classification,'' \emph{arXiv preprint
  arXiv:1709.00893}, 2017.

\bibitem{fan2018multi}
F.~Fan, Y.~Feng, and D.~Zhao, ``Multi-grained attention network for
  aspect-level sentiment classification,'' in \emph{Proceedings of the 2018
  conference on empirical methods in natural language processing}, 2018, pp.
  3433--3442.

\bibitem{huang2018aspect}
B.~Huang, Y.~Ou, and K.~M. Carley, ``Aspect level sentiment classification with
  attention-over-attention neural networks,'' in \emph{Social, Cultural, and
  Behavioral Modeling: 11th International Conference, SBP-BRiMS 2018,
  Washington, DC, USA, July 10-13, 2018, Proceedings 11}.\hskip 1em plus 0.5em
  minus 0.4em\relax Springer, 2018, pp. 197--206.

\bibitem{tang2016aspect}
D.~Tang, B.~Qin, and T.~Liu, ``Aspect level sentiment classification with deep
  memory network,'' \emph{arXiv preprint arXiv:1605.08900}, 2016.

\bibitem{song2019attentional}
Y.~Song, J.~Wang, T.~Jiang, Z.~Liu, and Y.~Rao, ``Attentional encoder network
  for targeted sentiment classification,'' \emph{arXiv preprint
  arXiv:1902.09314}, 2019.

\bibitem{xing2019earlier}
B.~Xing, L.~Liao, D.~Song, J.~Wang, F.~Zhang, Z.~Wang, and H.~Huang, ``Earlier
  attention? aspect-aware lstm for aspect-based sentiment analysis,''
  \emph{arXiv preprint arXiv:1905.07719}, 2019.

\bibitem{zhang2019syntax}
C.~Zhang, Q.~Li, and D.~Song, ``Syntax-aware aspect-level sentiment
  classification with proximity-weighted convolution network,'' in
  \emph{Proceedings of the 42nd international ACM SIGIR conference on research
  and development in information retrieval}, 2019, pp. 1145--1148.

\bibitem{nguyen2018effective}
H.~T. Nguyen and M.~Le~Nguyen, ``Effective attention networks for aspect-level
  sentiment classification,'' in \emph{2018 10th International Conference on
  Knowledge and Systems Engineering (KSE)}.\hskip 1em plus 0.5em minus
  0.4em\relax IEEE, 2018, pp. 25--30.

\bibitem{zhang2019aspect}
C.~Zhang, Q.~Li, and D.~Song, ``Aspect-based sentiment classification with
  aspect-specific graph convolutional networks,'' \emph{arXiv preprint
  arXiv:1909.03477}, 2019.

\bibitem{liang2020jointly}
B.~Liang, R.~Yin, L.~Gui, J.~Du, and R.~Xu, ``Jointly learning aspect-focused
  and inter-aspect relations with graph convolutional networks for aspect
  sentiment analysis,'' in \emph{Proceedings of the 28th international
  conference on computational linguistics}, 2020, pp. 150--161.

\bibitem{huang2019syntax}
B.~Huang and K.~M. Carley, ``Syntax-aware aspect level sentiment classification
  with graph attention networks,'' \emph{arXiv preprint arXiv:1909.02606},
  2019.

\bibitem{dong2014adaptive}
L.~Dong, F.~Wei, C.~Tan, D.~Tang, M.~Zhou, and K.~Xu, ``Adaptive recursive
  neural network for target-dependent twitter sentiment classification,'' in
  \emph{Proceedings of the 52nd annual meeting of the association for
  computational linguistics (volume 2: Short papers)}, 2014, pp. 49--54.

\bibitem{pontiki2014semeval}
M.~Pontiki, H.~Papageorgiou, D.~Galanis, I.~Androutsopoulos, J.~Pavlopoulos,
  and S.~Manandhar, ``Semeval-2014 task 4: Aspect based sentiment analysis,''
  \emph{SemEval 2014}, p.~27, 2014.

\bibitem{jiang2019challenge}
Q.~Jiang, L.~Chen, R.~Xu, X.~Ao, and M.~Yang, ``A challenge dataset and
  effective models for aspect-based sentiment analysis,'' in \emph{Proceedings
  of the 2019 conference on empirical methods in natural language processing
  and the 9th international joint conference on natural language processing
  (EMNLP-IJCNLP)}, 2019, pp. 6280--6285.

\bibitem{duchi2011adaptive}
J.~Duchi, E.~Hazan, and Y.~Singer, ``Adaptive subgradient methods for online
  learning and stochastic optimization.'' \emph{Journal of machine learning
  research}, vol.~12, no.~7, 2011.

\bibitem{loshchilov2017decoupled}
I.~Loshchilov and F.~Hutter, ``Decoupled weight decay regularization,''
  \emph{arXiv preprint arXiv:1711.05101}, 2017.

\bibitem{chen2017recurrent}
P.~Chen, Z.~Sun, L.~Bing, and W.~Yang, ``Recurrent attention network on memory
  for aspect sentiment analysis,'' in \emph{Proceedings of the 2017 conference
  on empirical methods in natural language processing}, 2017, pp. 452--461.

\bibitem{chen2020inducing}
C.~Chen, Z.~Teng, and Y.~Zhang, ``Inducing target-specific latent structures
  for aspect sentiment classification,'' in \emph{Proceedings of the 2020
  conference on empirical methods in natural language processing (EMNLP)},
  2020, pp. 5596--5607.

\bibitem{wang2020relational}
K.~Wang, W.~Shen, Y.~Yang, X.~Quan, and R.~Wang, ``Relational graph attention
  network for aspect-based sentiment analysis,'' \emph{arXiv preprint
  arXiv:2004.12362}, 2020.

\end{thebibliography}

\end{document}